\documentclass{article}

\usepackage[nonatbib, preprint]{neurips_2019}

\usepackage[utf8]{inputenc} 
\usepackage[T1]{fontenc}    
\usepackage{hyperref}       
\usepackage{url}            
\usepackage{booktabs}       
\usepackage{amsfonts}       
\usepackage{nicefrac}       
\usepackage{microtype}      
\usepackage{xcolor}
\usepackage{amsthm}
\usepackage{amsmath}
\usepackage{amssymb}
\usepackage{graphicx} 

\usepackage[numbers]{natbib}

\newtheorem{theorem}{Theorem}
\newcommand{\figref}[1]{\figurename~\ref{#1}}

\DeclareMathOperator{\divergence}{div}

\makeatletter
\newcommand{\printfnsymbol}[1]{%
  \textsuperscript{\@fnsymbol{#1}}%
}
\makeatother

\title{Equivariant Flows: sampling configurations for multi-body systems with symmetric energies}

\author{%
  Jonas Köhler \printfnsymbol{1}\printfnsymbol{2} \\
  \And 
  Leon Klein \printfnsymbol{1}\printfnsymbol{2} \\
  \And 
  Frank Noé \printfnsymbol{2}\printfnsymbol{3}\printfnsymbol{4}\\
  \AND
  \vspace{-2em}
  \\
  \printfnsymbol{2} Freie Universität Berlin, Department of Mathematics and Computer Science.\\
  \printfnsymbol{3} Freie Universität Berlin, Department of Physics.\\
  \printfnsymbol{4} Rice University, Department of Chemistry.\\
  \texttt{\{jonas.koehler, leon.klein, frank.noe\}@fu-berlin.de}\\
  \printfnsymbol{1} Authors contributed equally to this work.\\
}

\begin{document}

\maketitle

\begin{abstract}
Flows are exact-likelihood generative neural networks that transform samples from a simple prior distribution to the samples of the probability distribution of interest. Boltzmann Generators (BG) combine flows and statistical mechanics to sample equilibrium states of strongly interacting many-body systems such as proteins with 1000 atoms. In order to scale and generalize these results, it is essential that the natural symmetries of the probability density -- in physics defined by the invariances of the energy function -- are built into the flow. Here we develop theoretical tools for constructing such equivariant flows and demonstrate that a BG that is equivariant with respect to rotations and particle permutations can generalize to sampling nontrivially new configurations where a nonequivariant BG cannot.
\end{abstract}

\section{Introduction}

Generative learning using exact-likelihood methods based on invertible transformations has had remarkable success in domains like accurately representing images \cite{dinh2016density, Kingma_Glow, ho2019flow++}, audio \cite{VanDenOord_WaveNet2}, 3D point cloud data \cite{LiuQiGuibas_FlowNet3D,noe2019boltzmann} as well as applications in physics \cite{sharir2019deep, albergo2019flow, noe2019boltzmann}.

Recently, \textit{Boltzmann Generators} (BG) \cite{noe2019boltzmann} have been introduced for sampling Boltzmann type distributions $p(x) \propto \exp(-u(x))$ of high-dimensional multi-body problems, such as valid conformations of proteins. In contrast to typical generative learning problems, the target density $p(x)$ is specified by definition of the multi-body energy function $u(x)$ and the difficulty lies is learning to sample it efficiently. BGs do that in two steps: (1) An exact-likelihood method that generates samples from a density $q_{X, \theta}(x)$ that approximates the Boltzmann density $p(x)$. (2) An algorithm to reweigh the generated density to the target density $p(x)$.

In \cite{noe2019boltzmann}, it has been demonstrated that such BGs can be trained to efficiently sample prototypical many-body systems such as proteins in implicit solvent and condensed matter systems. In order to make further progress, it is essential to develop exact-likelihood generative models that respect the symmetries of $u(x)$, for example invariance of the energy with respect to global rotation or permutations of identical particles. 

In this work, we derive a sufficient criterion to design symmetry invariant BGs and design a simple instance that is symmetric with respect to translations, rotations and permutations. By comparing this model with the non-symmetric approach based on \textit{RealNVP} layers \cite{dinh2016density} as used in \cite{noe2019boltzmann}, we show that encoding symmetries in the model is critical for generalization.

\section{Boltzmann-generating flows}\label{boltzmann-generating-flows}

In order to reweigh the generated density $q_{X, \theta}(x)$ to the target density $p_{X}(x) \propto \exp(-u(x))$ (step 2 above), BGs require an exact likelihood generative model. This can be achieved by  transforming a simple prior density $q_{Z}(z)$, e.g. a multivariate Normal distribution, via an invertible function $f_{\theta}$ \cite{DinhDruegerBengio_NICE2015,dinh2016density, RezendeEtAl_NormalizingFlows, papamakarios2017masked,kingma2016improved,BehrmannEtAl_InvertibleResNets, grathwohl2018ffjord, durkan2019neural}. Sampling from $q_{X, \theta}(x)$ is achieved by sampling $z \sim q_{Z}(z)$ and transforming to $f_{\theta}(z) \sim q_{X, \theta}(x)$. Due to the invertibility of $f_{\theta}$, the probability density of any generated point can be computed with the \textit{change of variables} equation $$q_{X, \theta}(x) = q_{Z}\left(f_{\theta}^{-1}(x)\right) \det \frac{\partial f_{\theta}^{-1}(x)}{\partial x}.$$

Two obvious options are available in order to train the generator to match $q_{X, \theta} \approx p_{X}$ \citep{albergo2019flow, noe2019boltzmann}:
\begin{enumerate}
\item \textit{ML-training}: If some data $\left\{x_{n}\right\}_{n=1 \ldots N}$ is given that at least represents one or a few high-probability modes of $p(x)$, we can maximize the likelihood under the model, as is typically done when training flow-based models for images. 
\item \textit{KL-training} We minimize the reverse Kullback-Leibler divergence $KL(q_{X, \theta}\|p_{X})$. This approach is also known as energy-based training where the energy corresponding to the generated density is matched with $u(x)$. 
\end{enumerate}

\section{Equivariant Flows and Boltzmann Generators}
\subsection{Equivariant normalizing flows yield symmetric densities}

Symmetries may be discussed in terms of a \textit{group} $G$ acting on a vector space $V$. A \textit{representation} $\rho$ of $G$ is a map $\rho \colon G \rightarrow GL(V)$ satisfying $\rho(g)\rho(h) = \rho(g h)$. 
Examples are permutations and rotations in 3D which can be represented by permutation and rotation matrices.

We call any map $f \colon V \rightarrow V'$ $G$-\textit{invariant}, iff $f(\rho(g) x) = f(x)$ for all $g$ and $x$. We further call a map $f \colon V \rightarrow V$ $G$-\textit{equivariant}, iff $f(\rho(g) x) = \rho(g) f(x)$ for all $g$ and $x$.
In that sense an energy $u\colon \mathbb{R}^{n} \rightarrow \mathbb{R}$ is invariant with respect to a symmetry, if there is a symmetry group $G$, a representation $\rho$ of the group in the conformation space and the energy satisfies $u(\rho(g) x) = u(x)$.

Our first result is the following theorem, which gives a sufficient criterion, for when a transformed density resulting from a normalizing flow is symmetric

\begin{theorem}\label{thm:sufficient-criterion-classic-flows}
    Let $q_Z(z)$ be a $G$-invariant prior density and $f_{\theta}$ be a $G$-equivariant bijection. Let $q_{X, \theta}(x)$ be the density of $x = f(z)$ for $z \sim q_{Z}(z)$. Then $q_{X, \theta}$ is $G$-invariant.
\end{theorem}

\subsection{Equivariant dynamics yield equivariant continuous normalizing flows}

A recently proposed approach \cite{chen2018neural} to flow based modeling  describes the bijection implicitly via solving the Cauchy problem 
\begin{align}\label{eq:neural-ode}
    \frac{\partial y(t)}{\partial t} = g_{\theta}(y(t), t), \qquad y(t_0) = z.
\end{align}

for a fixed time horizon $[t_0, t_1]$, where $g_{\theta}$ can be a freely chosen / learned dynamics function. The solution of this initial value problem is $x := y(t_1) = y(t_0) + \int_{t_0}^{t_1} dt  g_{\theta}(y(t), t)$, which (under mild constraints) gives a bijection $z \rightarrow x$

As shown in \cite{chen2018neural}, the change of density can be computed at any integration time, using the \textit{continuous change of variable} rule, given by the differential equation
\begin{align}\label{eq:neural-ode-density}
    \frac{\partial \log p(y(t))}{\partial t} &= - \text{tr}\left[ \frac{\partial g_{\theta}(y(t), t)}{\partial y(t)}\right] = -\divergence g_{\theta}(y(t), t),\\ \nonumber \log p(y(t_0)) &= \log p(x).
\end{align}

Integrating eqs. \eqref{eq:neural-ode} \& \eqref{eq:neural-ode-density}, e.g. using a black-box solver, yields an exact-likelihood generative model.

For such \emph{continuous normalizing flows} (CNF), a statement similar to theorem \ref{thm:sufficient-criterion-classic-flows} can be made. Our second result is:
\begin{theorem}\label{thm:sufficient-criterion-continuous-flow}
    Let $f$ be a $G$-equivariant dynamics function. If $q_{Z}(z)$ is a $G$-invariant prior density, $z$ is sampled from $q_{Z}(z)$ and $x$ is obtained from $z$ by solving \eqref{eq:neural-ode}. Then $x$ is distributed according to a $G$-invariant density.
\end{theorem}


\subsection{Modeling symmetric multi-body systems with normalizing flows}

In this work we concentrate on modeling the equilibrium distribution of multi-body systems of $K$ particles ${x = (x_{1}, \ldots, x_{K}) \in \mathbb{R}^{K \times D}}$ ($D=2,3$), interacting with each other via a potential energy function $u(x_{1}, \ldots, x_{K})$. The equilibrium distribution of such a system is given by a Boltzmann-type distribution $p(x) \propto \exp\left(-u(x)\right)$ with an a priori defined energy $u(x)$. $u(x)$ is typically invariant with respect to certain transformations, here we assume the invariances of a molecule described by quantum mechanics without external field: 
\begin{enumerate}
\item Permutation invariance: Swapping the labels of any two interchangeable particles $(x_{1}, \ldots, x_{K}) \rightarrow (x_{\sigma(1)}, \ldots, x_{\sigma(K)})$.
\item Rotation invariance: Any 2D/3D rotation of the system $R x = (R x_{1}, \ldots, R x_{K})$.
\item Translation invariance: Any 2D/3D translation $x + v = (x_{1} + v, \ldots, x_{K} + v)$. 
\end{enumerate}

A natural prior distribution $q_{Z}(z)$ being invariant to all these symmetries and thus satisfying the conditions of thm. \ref{thm:sufficient-criterion-classic-flows} can be obtained by sampling ${z' = (z'_{1}, \ldots, z'_{N}) \sim \mathcal{N}(0, I_{D \times D})}$ i.i.d. and subtract its center of mass $\mu(x) := \sum_{i=1}^K x_{i}$ yielding $z = z' - \mu(z')$. For a mean-free configuration $z$, we can evaluate its likelihood by $\log q_{Z}(z) = -1/2 \sum_{i}^{K} \|z_{i}\|^2 + const.$, which is invariant with respect to the symmetries 1.-3. above.



\subsection{Equivariant flows for multi-body systems}

We give a simple example of an equivariant flow which obeys the symmetries 1.-3. above. For this, we employ thms. \ref{thm:sufficient-criterion-classic-flows} \& \ref{thm:sufficient-criterion-continuous-flow} to design equivariant dynamics functions with tractable exact traces.

Each particle is updated using a vector field in radial direction
\begin{align}
    \label{eq:equivariant-dynamics-radial}
    \frac{\partial x_{i}(t)}{\partial t} = g_{\theta}(x(t), t)_{i} := \sum_{j = 1}^{K} \psi_{\theta}\left(\| x_{i} - x_{j} \|_{2}\right) \left( x_{i} - x_{j} \right)
\end{align}

where $\psi_{\theta} \colon \mathbb{R}_{\geq 0} \rightarrow \mathbb{R}$ can be an arbitrary scalar function. These dynamics correspond to particle updates due to a conservative force generated by a potential that depends on particle distances. The divergence is then given by
\begin{align}
    \label{eq:equivariant-dynamics-divergence}
    \divergence g_{\theta}(x(t), t) = \sum_{i=1}^{K}\sum_{j=1}^{K} \frac{\partial \psi_{\theta}(\| x_{i} - x_{j}\|_{2})}{\partial \| x_{i} - x_{j}\|_{2}} \| x_{i} - x_{j}\|_{2} + D \sum_{i=1}^{K}\sum_{j=1}^{K} \psi_{\theta}(\| x_{i} - x_{j}\|_{2}),
\end{align}
which can be computed exactly in an efficient way using one backpropagation pass through the graph.


It is simple to see that this vector field is equivariant with respect to symmetries 1.-3.. Further this field does not alter the center of mass of a particle system, thus integrating mean-free states along it will always result in mean-free states.

\section{Experiments}

\begin{figure}
  \centering
  
  \includegraphics[width=\textwidth]{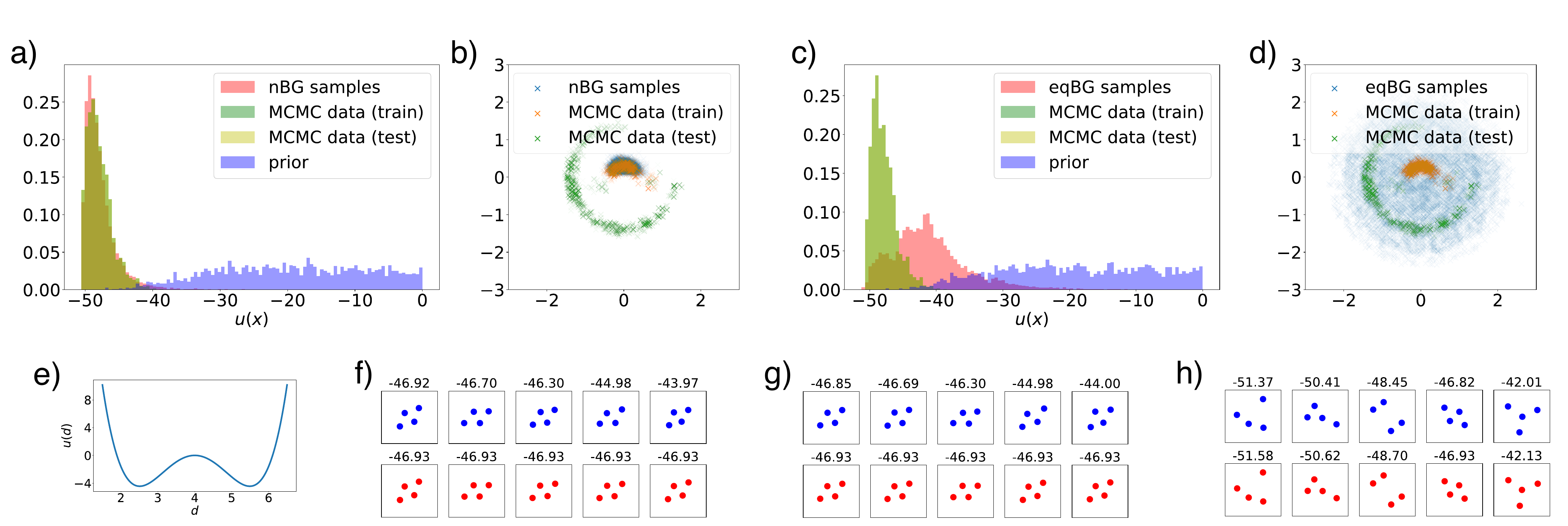}
  \vspace{-0.5cm}
  \caption{\textbf{Comparison: sampling equivariant and non-equivariant Boltzmann Generators}. \textbf{a, c)} Energies for samples taken from trained nBG and eqBG respectively. \textbf{b, d)} Marginal distribution of the first particle position $x_1$ for states sampled from trained nBG and eqBG respectively. \textbf{e)} Potential energy of the system illustrated for $N=2$. \textbf{f-h)} \textit{Top row:} sampled states from training data, trained nBG and trained eqBG respectively together with corresponding energies. \textit{Bottom row:} corresponding configurations after energy minimization.}
  \label{fig:big-figure}
\end{figure}

Using a simple 2D toy system, we show empirically, that a Boltzmann Generator using an equivariant flow encoding the symmetries of the system's energy function explicitly as given in the last section (\textit{eqBG}) can generalize beyond the training data whereas a non-equivariant approach (furthermore abbreviated as \textit{nBG}) based on RealNVP transformations cannot.


Our system ($N=4, D=2$) is given by the pairwise potential 
$${u(x) = \sum\limits_{i=1,j=1, j\neq i}^{N}  -4 \left(\|x_{i} - x_{j} \|_{2} - 4\right)^{2} + 0.9 \left(\|x_{i} - x_{j} \|_{2} - 4\right)^{4}},$$
which produces two distinct low energy modes separated by a high energy barrier (see \figref{fig:big-figure} e). By coupling multiple particles with such double-well interactions we can create a frustrated system with multiple metastable states.

\subsection{First experiment: generalization on unseen trajectories}

In a first experiment, we compare how both models generalize to unseen trajectories. For this we use standard MCMC sampling using the classic Metropolis-Hasting algorithm to produce short out-of-equilibrium trajectories of the system. We split a trajectory using the first 50\% of samples for initializing both, the nBG and the eqBG, with ML training. After that we fine-tune both models on the energy using a weighted sum of ML and KL training. Both models are able to sample states with acceptable energies (\figref{fig:big-figure} a, c). However, evaluating both models on the second 50\% of the trajectory, we clearly see that the nBG is not generalizing at all, whereas the eqBG is able to achieve similar likelihoods on train and test set simultaneously (\figref{fig:big-figure} b, d, Table \ref{tab:likelihoods}). 

\begin{table}[h]
    \caption{Negative log-likelihood of trained equivariant and non-equivariant Boltzmann Generators.}
      \label{table}
    \centering
    \begin{tabular}{c c | c c}
        \toprule
         nBG (train)  & nBG (test) & eqBG (train) & eqBG (test) \\
            \midrule
         -12.64 & 372.69 & 5.91 & 6.72\\
    \end{tabular}
    \label{tab:likelihoods}
\end{table}

\subsection{Second experiment: discovery of unseen metastable states}

In a second experiment, we evaluate to which extend both models are able to discover new metastable states, which have not been observed in the data set. Here we (1) create a training set, by perturbing a single minimum state with a tiny amount of Gaussian noise (\figref{fig:big-figure} f), (2) initialize both, the nBG and eqBG, with ML training on the training set, (3) fine-tune both models using a weighted sum ML and KL training on the energy, (4) generate samples from the converged models (\figref{fig:big-figure} g,h, top row) and (5) locally minimize the energies $u(x)$ of the generated structures $x$ using a non-momentum second-order optimizer (\figref{fig:big-figure} g, h, bottom row). As can be seen, the nBG will only reproduce the unique state from which the training set was constructed. However, the eqBG will produce a variety of local minima, which did not appear in the training set.

\section{Discussion}
We gave a sufficient criterion, by which equivariant flows can be constructed and implemented an instance for symmetries frequently appearing in multi-particle systems. Using a simple example we have demonstrated that encoding symmetries of the target density/energy into the invertible function is critical, if we expect generative flows such as BGs to generalize beyond seen trajectories. Scaling such equivariant flows and BGs to large real-world systems remains a major challenge for future research.



\appendix

\end{document}